# The Transfer of Evolved Artificial Immune System Behaviours between Small and Large Scale Robotic Platforms

Amanda M. Whitbrook, Uwe Aickelin, and Jonathan M. Garibaldi

Intelligent Modelling and Analysis Research Group (IMA),
School of Computer Science, University of Nottingham, Nottingham, NG8 1BB
amw@cs.nott.ac.uk, uxa@cs.nott.ac.uk, jmg@cs.nott.ac.uk

**Abstract.** This paper demonstrates that a set of behaviours evolved in simulation on a miniature robot (epuck) can be transferred to a much larger scale platform (a virtual Pioneer P3-DX) that also differs in shape, sensor type, sensor configuration and programming interface. The chosen architecture uses a reinforcement learning-assisted genetic algorithm to evolve the epuck behaviours, which are encoded as a genetic sequence. This sequence is then used by the Pioneers as part of an adaptive, id- iotypic artificial immune system (AIS) control architecture. Testing in three different simulated worlds shows that the Pioneer can use these behaviours to navigate and solve object-tracking tasks successfully, as long as its adaptive AIS mechanism is in place.

## 1 Introduction

Evolutionary robotics is a technique that refers to the genetic encoding of autonomous robot control systems and their improvement by artificial evolution. Ideally the end product should be a controller that evolves rapidly and has the properties of robustness, scalability and adaptation. However, in practice it proves difficult to achieve all of these goals without introducing an additional mechanism for adaptability since behaviour is essentially an emergent property of interaction with the environment [11]. Thus, the major challenge facing evo- lutionary robotics is the development of solutions to the problem of brittleness via the design of controllers that can generalize to modified environments.

The characteristics of the robot body and its sensorimotor system may be regarded as part of the environment [2] as all embodied systems are physically embedded within their ecological niche and have a dynamic reciprocal coupling to it [9]. Indeed, artificial evolution often produces control systems that rely heavily on body morphology and sensorimotor interaction [3], and when these are subsequently altered, the changes can affect behavioural dynamics drasti- cally. Thus, a solid test for robustness, scalability and adaptation is the ability of an evolved control system to function not only in different physical environ- ments, but also on a number of robotic platforms that differ in size, morphology, sensor type and sensor-response profile. This paper is therefore concerned with demonstrating the theoretical and practical cross platform transferability of an evolutionary architecture designed to combat adaptation problems.

Adaptation is usually made possible through the introduction of additional mechanisms that permit some kind of post-evolutionary behaviour modification. The architecture used here falls into the general category of approaches that combine evolution or long term learning (LTL) with a form of lifelong or short term learning (STL) to achieve this [14]. The particular technique consists of the rapid evolution of a number of diverse behaviour sets using a Webots [10] simulation (LTL) followed by the use of an idiotypic artificial immune system (AIS) for selecting appropriate evolved behaviours as the robot solves its task in real time (STL). The approach differs from most of the evolutionary schemes employed previously in the literature in that these are usually based on the evolution of neural controllers [6] rather than the actual behaviours themselves.

Previous papers have provided evidence that the idiotypic AIS architecture has advantages over a reinforcement learning scheme when applied to mobile robot navigation problems [15] and have shown that the idiotypic LTL-STL architecture permits transference from the simulator to a number of different real-world environments [16]. The chief aim of this paper is, therefore, to supply further support for the robustness, scalability and adaptability of the architec- ture by showing that it can be extended to the much larger scale Pioneer P3-DX robots. For this purpose, the behaviours evolved for the epuck in the Webots simulator are transplanted onto the Pioneer and used both with and without the idiotypic network in Player's Stage [5] simulator. The results successfully demonstrate the scalability of the method in the virtual domain, and provide strong empirical evidence that the idiotypic selection feature is a vital component for achieving this.

The remainder of this paper is structured as follows. Section 2 introduces some essential background information about the problem of platform transfer in mobile robotics and previous attempts to achieve it. Section 3 describes the LTL-STL control system, including its modular structure and the encoding of the evolved behaviours. In particular, it shows how platform transfer is achieved between the epuck and Pioneer P3-DX robots. Section 4 provides information regarding the simulated test environments and experimental set-up and Section 5 presents and discusses the results. Section 6 concludes the paper.

## 2   Background and Relevance

Cross platform transfer of an intelligent robot-control algorithm is a highly desirable property since more generic software is more marketable for vendors and more practical for users with more than one robot type. Furthermore, software that is robust to changes in a user's hardware requirements is particularly at- tractive. However, transferability between platforms is difficult to achieve and is hence extremely rare in mobile robotics [2]. This is primarily due to hardware differences such as the size, morphology and spatial relationships between the body, actuators and sensors, which constitute a drastic change of the environ- ment from an ecological perspective. Since modern mobile robot systems are distributed systems, transfer may also be hindered by diversity of middleware, operating systems, communications protocols and programming languages and their libraries [13]. Furthermore, portability is made even more challenging by differences in sensor type, sensor characteristics and the mechanical structure of the robot.

Despite its rarity, platform transfer for evolved control systems is reported in the literature. Floreano and Mondada [2,3] use an incremental approach to evolve artificial neural networks for solving a looping maze navigation problem. Evolution begins with a real miniature Khepera robot and gradually moves to a Koala, a larger, more fragile robot supplied by the same manufacturer. Within this architecture, previously evolved networks are gradually adapted, combined, and extended to accommodate the changing morphology and sensorimotor inter- faces [3]. However, the scheme possesses some significant drawbacks. The use of physical robots for the evolution is both impractical and infeasible due to the ex- cessive amount of time and resources required. For example, adaptability to the Koala platform emerges only after 106 generations on the real Khepera and an additional 30 on the Koala (each generation taking approximately 40 minutes). Also, if each new environment or platform requires additional evolution then there is no controller that is immediately suitable for an unseen one. Another consideration is that the Koala was deliberately designed to support transfers from the Khepera and is thus very similar in terms of the wheel configuration, IR sensors, vision module, low-level BIOS software and communication method. This is good from a practical perspective, but one could argue that the new, supposedly unknown environment is far too engineered.

Floreano and Urzelai [4,12] also evolve a neural network to control a light-switching robot but evolve the mechanisms for self-organization of the synaptic weights rather than the weights themselves. This means that the robot is rapidly able to adapt its connection weights continuously and autonomously to achieve its goal. The authors transfer the evolved control system from simulation to a real Khepera and also from a simulated Khepera to a real Koala. They report no reduction in performance following the transfer. However, since the same platforms are used as in [2,3], the new environment is, again, too engineered. In addition, the task is very simple and the environment is sparse, requiring minimal navigational and obstacle avoidance skills.

In this paper more complex tasks and environments are used to demonstrate that behaviours evolved on a simulated epuck can be used by a larger, unrelated robot that has not deliberately been designed for ease of transfer (the Pioneer P3-DX). This represents a much more difficult platform transfer exercise than has been attempted before and is hence a more realistic test of control system adaptability. In particular, Pioneer P3-DX robots differ significantly from epucks in mechanical structure, body size, body shape and wheel size and possess sixteen sonar sensors rather than the eight infrared (IR) sensors of the epuck, which also have a different spatial arrangement. The Pioneer is also produced by a different manufacturer and uses different middleware and a different simulator (Stage [5]). A full comparison between the two platforms is provided in Section 3.4, Table 1. In addition, the achievement of platform transfer between epucks and Pioneers is of practical value since Pioneer behaviours cannot be evolved directly on the Stage simulator within a realistic time frame; Stage is not fast or accurate enough, and control systems used in the Webots programming environment are not directly transferable to real Pioneers. Moreover, simulation of the epuck in Webots requires a 3D model that is readily available, so it is computationally much cheaper to reuse the epuck's evolved behaviours in Stage rather than to design a complex 3D Pioneer model for Webots.

## 3  System Architecture

### 3.1  Artificial Immune Systems and the Behavioural Encoding

AISs mimic the properties of the vertebrate immune system (for example antibody recognition of and stimulation by antigens). Idiotypic systems in particular exploit Jerne's notion of an idiotypic network [7], where antibodies are capable of recognizing and being stimulated or suppressed by other antibodies via their paratopes and idiotopes, see [15] for further details. Idiotypic AIS algorithms are often based on Farmer *et al.*'s computational model [1] of Jerne's theory and are characterized by a decentralized behaviour-selection mechanism. They are a popular choice for STL in robotic control systems [8] since they allow much greater flexibility for determining a robot's actions. The LTL and STL aspects of the control system presented here thus work together to produce adaptability; diversity of the behaviour sets is provided by the evolutionary (LTL) component and the idiotypic network (STL) exploits the wide range of choice available to select behaviours appropriate for a given environmental scenario.

The AIS analogy is that antigens model the environmental information as perceived by the sensors and antibodies model the behaviours of the robot. Here, eight antigens (coded 1 - 8) are identified based on the robot's possession of distance-measuring sensors (IR or sonar) and a camera for tracking coloured objects. These are 1 - target unseen, 2 - target seen, 3 - obstacle right, 4 - obstacle rear, 5 - obstacle left, 6 - collision right, 7 - collision rear, 8 - collision left. In addition, six types of basic behaviour (coded 1 - 6) are used; 1 - wandering using either a left or right turn, 2 - wandering using both left and right turns,
3 - turning forwards, 4 - turning on the spot, 5 - turning backwards, and 6 - tracking targets. More detailed individual behaviours are thus described using the attribute type $T$, which refers to the basic behaviour code, and the additional attributes speed $S$ in epuck speed units per second ($\psi$ per second), frequency of turn $F$ (% of time), angle of turn $A$ (% reduction in one wheel speed), direction of turn $D$ (either 1 - left or 2 - right), frequency of right turn $R_f$ (% of time) and angle of right turn $R_a$ (% reduction in right wheel speed). This structure means that, potentially, a vast number of diverse behaviours can be created. However, there are limits to the attribute values [17]; these are carefully selected in order to strike a balance between reducing the size of the search space, which increases speed of convergence, and maintaining diversity. More details on the behavioural encoding are provided in Section 3.4.

### 3.2   LTL Phase

The LTL phase is a reinforcement learning-assisted genetic algorithm (GA) that evolves a suitable behaviour for each antigen. It works by selecting two differ- ent parent robots via the roulette-wheel method and determining behaviour at- tribute values for their offspring as described in [17]. The reinforcement learning component constantly assesses the performance of the behaviours during evo- lution so that poorly-matched ones are replaced with newly-created ones when the need arises, which accelerates the GA. All the test problems are assessed by measuring task completion time $t_i$ and number of collisions $c_i$ $i = 1, ..., x$, thus, the relative fitness $\mu_i$ of each member of the population is calculated using:

$$\mu_i = \frac{1}{t_i + \rho c_i \sum_{k=1}^{x} (t_k + \rho c_k)^{-1}}, \qquad (1)$$

where $\rho$ represents the weighting given to collisions ($\rho = 1$ here) and $x$ is the number of robots in the population. After convergence, the fittest robot in the final population is selected. However, since the idiotypic network requires a number $n$ of distinct behaviours for each antigen, the whole process is repeated $n$ times in order to obtain $n$ robots from separate populations that never interbreed. This is an alternative to selecting a number of robots from a single, final population and means that greater diversity can be achieved with smaller population sizes without reducing speed of convergence. The attribute values representing the behaviours of the $n$ robots and their final reinforcement scores are saved as a genetic sequence (a simple text file) for seeding the AIS system.

### 3.3   STL Phase

The AIS system reads the genetic sequence generated by the LTL phase and then calculates the relative fitness $\mu_i$ of each behaviour (or antibody) set using (1), where $\rho = 8$ to increase the weighting given to the number of collisions. It then produces an $n \times 8$ matrix $P$ (analogous to an antibody paratope) representing the reinforcement scores, or degree of match between antibodies and antigens. The elements of this matrix ($P_{ij}$ $i = 1, ..., n, j = 1, ..., 8$) are calculated by multiplying each antibody's final reinforcement score by the relative fitness of its set $\mu_i$. An $n \times 8$ matrix $I$ (analogous to an antibody idiotope) is also created by assigning a value of 1.0 to the element corresponding to the minimum $P_{ij}$ for each $j$, and designating a value of 0.0 to all other elements. The matrix $P$ is adjusted after every iteration through reinforcement learning, but $I$ remains fixed throughout. If an idiotypic network is not used then $P$ alone governs antibody selection; the antigen-matching antibody with the highest reinforcement score is used. If idiotypic stimulation and suppression of antibodies are taken into account then $I$ is used to adjust the degree of match of each antibody as described in [16], which may result in a different one being selected.

### 3.4 Mechanisms of Cross Platform Transfer

Following convergence of the GA, the selected behaviours are encoded as nine integers in a simple text file that contains all the genetic information necessary to reproduce them. The first integer represents the antigen code, and the next seven represent the behavioural attributes $T$, $S$, $F$, $A$, $D$, $R_f$ and $R_a$. The last integer is the final reinforcement score attained by the behaviour prior to convergence. The genetic sequence encodes the principal wheel speeds in epuck speed units per second ($\psi$ per second) where $\psi = 0.00683$ radians. A speed value of 600 $\psi$ per second thus corresponds to $600 \times 0.00683 = 4.098$ radians per second. An example line from a genetic text file is: 0 2 537 80 51 2 37 76 50, which encodes wandering in both directions with a speed of 537 $\psi$ per second, turning 80% of the time. The robot turns right 37% of this time by reducing the speed of the right wheel by 76%, and turns left 63% of this time by reducing the speed of the left wheel by 51%. A particular genetic sequence thus governs how the left and right wheel speeds change with time.

In theory, the behavioural encoding may be extended to any two-wheeled, non-holonomic, mobile-robot, since the wheel motions of such robots are fully described by their changing speeds. Furthermore, since the output from the LTL phase is a simple text file, any program is capable of reading it and extracting the information necessary to form the wheel motions. Moreover, specification of the speeds in radians per second permits automatic scaling between different-sized environments, without requiring knowledge of the particular scales involved, since wheel size is generally related to the scale of the environment. However, it is also necessary to consider some fundamental hardware and software differences between the Pioneer P3-DX and epuck robots when making the transfer. Table 1 below shows the technical specification for each robot type. The most fundamental considerations are the larger scale of the Pioneer, the use of different programming environments, the use of sonar sensors on the Pioneer and the spatial arrangement of these sensors. These affect the transfer in three main ways; how velocity is expressed, how the sensors are read and translated into antigen codes, and how blob finding is implemented.

Use of the genetic sequence coupled with a simple conversion of $\psi$ per second to radians per second, as described above, would be adequate to cater for the scaling differences if the two platforms did not use different APIs. However, the epuck is programmed using the Webots C/C++ Controller API, where robot wheel speeds are set using the differential wheels set speed method, which requires the left and right wheel speeds in $\psi$ per second as its arguments. In contrast, the Pioneer robot is programmed using libplayerc++, a C++ client library for the Player server. In this library, the angular and linear components of the robot's velocity are set separately using yaw $\omega$ and velocity $v$ arguments for the SetSpeed method of the Position2dProxy class, and $v$ is expressed in metres per second. As methods for encoding the genetic sequence into left $L$ and right $R$ epuck wheel speeds already exist, it is computationally cheaper to reuse these methods on the Pioneer and simply convert them into equivalent $\omega$ and $v$ arguments. The conversions are given by:

Table 1. Differences between the Pioneer and Epuck Robotic Platforms

| No. | Attribute | Pioneer P3-DX | Epuck |
| --- | --- | --- | --- |
| 1 | Manufacturer | MobileRobots Inc | EPFL |
| 2 | Simulator | Stage | Webots |
| 3 | Middleware | Player | Webots |
| 4 | Operating system | Linux | N/A |
| 5 | Communications protocol | Wireless TCP/IP | Bluetooth |
| 6 | Wheel radius (cm) | 9.50 | 2.05 |
| 7 | Wheel width (cm) | 5.00 | 0.20 |
| 8 | Axle length (cm) | 33.00 | 5.20 |
| 9 | Body material | Aluminium | Plastic |
| 10 | Body length (cm) | 44 | 7 |
| 11 | Body width (cm) | 38 | 7 |
| 12 | Body height (cm) | 22 | 4.8 |
| 13 | Weight (kg) | 9 | 0.15 |
| 14 | Body shape | Octagonal | Circular |
| 15 | Sensor type | Sonar | Infrared |
| 16 | No. of sensors | 16 | 8 |
| 17 | Sensor range | 15cm to 5m | 0 to 6cm |
| 18 | Camera | Canon VC-C4 | VGA |
| 19 | Blob finding software | Player | Weblobs |

$$v = \frac{\psi r_p (R + L)}{2}, \qquad (2)$$

$$\omega = \frac{\zeta \psi r_e (R - L)}{a_e}, \qquad (3)$$

where $r_p$ is the radius of the Pioneer wheel, $r_e$ is the radius of the epuck wheel, and $a_e$ is the axle length of the epuck. The parameter $\zeta = 1.575$ is determined by empirical observation and is introduced in order to replicate the angular movement of the epuck more accurately.

The antigens indexed 3 to 8 describe an obstacle's orientation with respect to the robot (right, left or rear) and classify its distance from the robot as either "obstacle" (avoidance is needed) or "collision" (escape is needed). Thus, two threshold values $\tau_1$ and $\tau_2$ are required to mark the boundaries between "no obstacle" and "obstacle" and between "obstacle" and "collision" respectively. The epuck's IR sensors are nonlinear and correspond to the quantity of reflected light, so higher readings mean closer obstacles. In contrast, the Pioneer's sonar readings are linear denoting the estimated proximity of an obstacle in metres, so lower readings mean closer obstacles. Since direct conversion is difficult, the threshold values $\tau_1$ and $\tau_2$ (250 and 2400 for the epuck) are determined for the Pioneer by empirical observation of navigation through cluttered environments, ($\tau_1 = 0.15$ and $\tau_2 = 0.04$). Additionally, in order to determine the orientation of any detected obstacle, the epuck uses the index of the maximum IR reading, where indices 0, 1 and 2 correspond to the right, 3 and 4 correspond to the rear and 5, 6 and 7 correspond to the left. For the Pioneer it is necessary to use the index of the minimum sonar reading and encode positions 4 to 9 as the right,
10 to 13 as the rear and positions 0 to 3 and 14 to 15 as the left, due to the different spatial arrangement of the sensors.

Blob finding software (named Weblobs) was developed for the epuck as part of this research, since the Webots C/C++ Controller API has no native blob finding methods. However, the Pioneer robot is able to use methods belonging to the BlobfinderProxy class of libplayerc++. The objective is to determine whether blobs (of the target colour) are visible, and if so, to establish the direction (left, centre or right) of the largest from the centre of the field of view. The two robot types thus use different blob finding software, but collect the same information.

### 3.5 Modular Control Structure

The entire STL program is broken down into the pseudo code below in order to demonstrate its modular structure and the ease with which this facilitates adaptation for the Pioneer P3-DX platform. Each block shows the module it calls and the method it uses within that module. Blocks marked with an asterisk are dealt with in the main body of the program and do not call other modules.

```
1     Initialize robot(ROBOT --> InitializeRobot() --> InitializeSensors())
2     Read genetic sequence *
3     Build matrices P and I *
      REPEAT
4         Read sensors (ROBOT --> ReadSensors())
5         Read camera (BLOBFINDER --> GetBlobInfo())
6         Determine antigen code *
7         Score previous behaviour using reinforcement learning
8         Update P *
9         Select behaviour *
10        Update antibody concentrations *
11        Execute behaviour (BEHAVIOUR --> Execute())
      UNTIL stopping criteria met
```

The only blocks that require changes for the Pioneer platform are 1, 4, and 5. Since these are dealt with by calling other modules, the main body of the program can be wholly reused, although an additional two lines in block 11 are necessary to convert the wheel speeds to the Player format. Some slight changes to block 7, which deals with using sensor data to determine the reinforcement score are also required.

## 4 Test Environments and Experimental Set-Up

The genetic behaviour sequences are evolved using 3D virtual epucks in the Webots simulator, where the robot is required to track blue markers in order to navigate through a number of rooms to the finish line, see [17]. Throughout evolution, five separate populations of ten robots are used and the mutation rate
$б$ is set at 5% as recommended by [17] for a good balance between maximizing diversity and minimizing convergence time. Following the LTL phase, the evolved behaviour sequences are used with 2D virtual Pioneer robots in three different Stage worlds, $S_1$, $S_2$, and $S_3$. $S_1$ and $S_2$ require maze navigation and the tracking of coloured door markers (Figure 1 and Figure 2), and $S_3$ involves search and retrieval of a blue block whilst navigating around other obstacles (Figure 3). Sixty runs are performed in each Stage world, thirty using the idiotypic selection mechanism, and thirty relying on reinforcement learning only. In addition, in $S_3$ the obstacle positions, target block location, and robot start point are changed following each idiotypic and nonidiotypic test, so that the data is paired. For all runs, the task time $t$ and number of collisions $c$ are recorded. However, a fast robot that continually crashes or a careful robot that takes too long to complete the task is undesirable, so an additional score metric $\varphi$ that combines $t$ and $c$ is computed for each run. This is given by:

$$\varphi = \frac{t + \sigma_i c}{2}, \quad (4)$$

where $\sigma_i$ is the ratio of the mean task time $\bar{t}$ to mean number of collisions $\bar{c}$ for world $S_i$. In all worlds, $\bar{t}$, $\bar{c}$ and $\bar{\varphi}$ are computed with and without using idiotypic effects and the results are compared using a 2-tailed t-test (paired for world $S_3$), with differences accepted as significant at the 99% level only. As another measure of task performance, runs with an above average $\varphi$ for each world are counted as good and those with fitness in the bottom 10% of all runs in each world are counted as bad. Additionally, for each task, robots taking longer than 900 seconds are counted as having failed and are stopped.

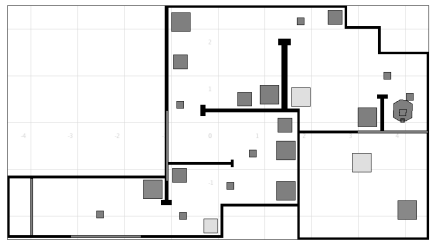

Fig. 1. 2D Stage world $S_1$ used in the Pioneer STL phase

## 5   Results and Discussion

Table 2 shows $\bar{t}$, $\bar{c}$ and $\bar{\varphi}$ values with and without using idiotypic effects in each world and the significant difference levels when these are compared. It also displays the percentage of good and bad runs and number of fails in each world.

When idiotypic effects are employed, the virtual Pioneer robots prove able to navigate safely (the mean number of collisions is very low) and solve their tasks within the alloted time in all of the worlds. Navigation is also safe for the nonidiotypic robots, but, in terms of task time in worlds $S_1$ and $S_2$ there is a 17% failure rate and in world $S_3$ there is a 7% failure rate. Furthermore, mean task time is significantly higher than for the idiotypic case in all worlds, although the number of collisions is consistently low and not significantly different between the idio- typic and nonidiotypic cases. In addition, the number of bad runs is higher and the number of good runs is lower for nonidiotypic robots in all worlds and the score is significantly better when idiotypic effects are employed in worlds $S_1$ and $S_2$. These observations provide strong empirical evidence that the behaviours evolved in simulation on an epuck robot can be successfully ported to the vir- tual Pioneer P3-DX platform provided that the adaptive idiotypic mechanism is applied within the STL architecture. As with the STL results for virtual and real epucks (documented in [16]), the results show that the evolutionary (LTL) phase is capable of producing sets of very diverse behaviours but that the STL phase requires the use of a scheme that selects from the available behaviours in a highly adaptive way. This is further illustrated in Figure 3 which shows the paths taken by an idiotypic (left) and nonidiotypic (right) Pioneer when solving the block-finding problem in world $S_3$. It is evident that the nonidiotypic Pioneer takes a much less direct route and repeats its path several times. This is because it is less able to adapt its behaviour and consequently spends much more time wandering, getting stuck and trying to free itself. This result is typical of the $S_3$ experiments.

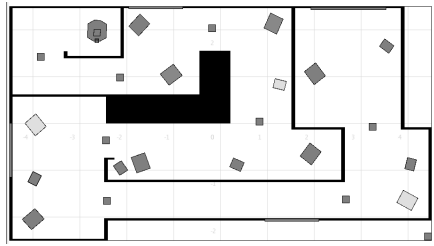

Fig. 2. 2D Stage world $S_2$ used in the Pioneer STL phase

The chosen architecture has a number of benefits. The reinforcement-assisted GA effectively balances the exploitative properties of reinforcement and the explorative properties of the GA. This reduces convergence time, improves accu-racy and maintains GA reliability. The genetic encoding of the behaviours and the choice of separate populations permits greater diversity for the antibodies, which allows for a much more adaptive strategy in the STL phase. Earlier work [15] suggests that the idiotypic advantage can be attributed to an increased rate of antibody change, which implies a much less greedy strategy. It also proposes that the network is capable of linking antibodies of similar type, so that useful but potentially untried ones can be used. The use of concentrations and feedback within the network may also facilitate a memory feature that achieves a good balance between selection based on past antibody use and current environmental information. However, the present scheme has some limitations; there is no scope to change the antibodies within the network, only to choose between them. A possible improvement would be constant execution of the LTL phase, which regularly updates the genetic sequence, allowing fresh antibodies to be used if the need arises. In addition, success with transference to other platforms is presently too heavily dependent upon parameter tuning and readjustment of the reinforcement scheme for the particular sensor characteristics.

Table 2. Results of Experiments with and without Idiotypic Effects in Each World. G = % Good, B = % Bad, F = % Fail.

| World | Significance | | | Idiotypic | | | | | | Nonidiotypic | | | | | |
|---|---|---|---|---|---|---|---|---|---|---|---|---|---|---|---|
| | $\bar{t}$ | $\bar{\tau}$ | $\bar{\varphi}$ | $\bar{t}(s)$ | $\bar{\tau}$ | $\bar{\varphi}$ | G | B | F | $\bar{t}(s)$ | $\bar{\tau}$ | $\bar{\varphi}$ | G | B | F |
| $S_1$ | 100 | 96 | 100 | 176 | 2 | 166 | 70 | 3 | 0 | 336 | 4 | 346 | 47 | 30 | 17 |
| $S_2$ | 100 | 98 | 100 | 309 | 2 | 287 | 27 | 10 | 0 | 513 | 5 | 535 | 13 | 60 | 17 |
| $S_3$ | 100 | 56 | 97 | 160 | 2 | 233 | 43 | 7 | 0 | 395 | 1 | 322 | 37 | 37 | 7 |

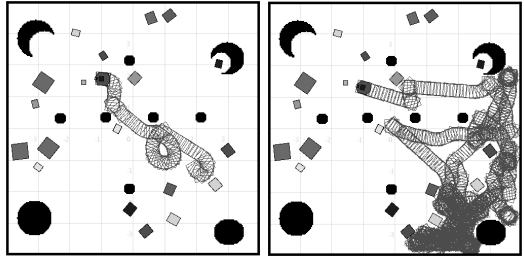

Fig. 3. World $S_3$ showing the trail of an idiotypic (left) and nonidiotypic (right) robot

## 6  Conclusions

This paper has described a mobile robot control architecture that consists of an LTL (evolutionary) phase responsible for the generation of sets of diverse be- haviours, and an STL (immune system) phase, which selects from the available behaviours in an adaptive way. It has shown that the behaviours are essentially platform independent and that they can be evolved in simulation on a miniature epuck robot and used on a much larger virtual Pioneer P3-DX robot. The plat- form transfer is equivalent to a complex and difficult environmental change and is thus a sound test of adaptability and scalability for the combined LTL-STL architecture. Tests in different environments have shown that the Pioneer is able to accomplish navigation, obstacle avoidance and retrieval tasks using the epuck behaviours, and that on average it performs significantly faster when employing the idiotypic mechanism as behaviour selection is much more adaptable than using reinforcement learning alone. The next step is testing with real Pioneer P3-DX robots to establish whether similar levels of success can also be achieved in the real domain.